\documentclass[fleqn,10pt]{wlscirep}
\usepackage{amssymb}
\usepackage{bbm}
\usepackage{amsmath}

\title{Multimodal and Multiscale Deep Neural Networks for the Early Diagnosis of Alzheimer's Disease using structural MR and FDG-PET images}

\author[1,*]{Donghuan Lu}
\author[1]{Karteek Popuri}
\author[1]{Weiguang Ding}
\author[1]{Rakesh Balachandar}
\author[1]{Mirza Faisal Beg}
\author[+]{for the Alzheimer’s Disease Neuroimaging Initiative}
\affil[1]{School of Engineering Science, Simon Fraser University, Burnaby, V5A 1S6, Canada}

\affil[*]{ludonghuan9@gmail.com}
\affil[+]{Data used in preparation of this article were obtained from the Alzheimer’s Disease Neuroimaging Initiative (ADNI) database (\url{adni.loni.usc.edu}). As such, the investigators within the ADNI contributed to the design and implementation of ADNI and/or provided data but did not participate in analysis or writing of this report. A complete listing of ADNI investigators can be found at: \url{http://adni.loni.usc.edu/wp-content/uploads/how_to_apply/ADNI_Acknowledgement_List.pdf}}


\begin{abstract}
Alzheimer's Disease (AD) is a progressive neurodegenerative disease.  Amnestic mild cognitive impairment (MCI) is a common first symptom before the conversion to clinical impairment where the individual becomes unable to perform activities of daily living independently. Although there is currently no treatment available, the earlier a conclusive diagnosis is made, the earlier the potential for interventions to delay or perhaps even prevent progression to full-blown AD. Neuroimaging scans acquired from MRI and metabolism images obtained by FDG-PET provide in-vivo view into the structure and function (glucose metabolism) of the living brain. It is hypothesized that combining different image modalities could better characterize the change of human brain and result in a more accuracy early diagnosis of AD. In this paper, we proposed a novel framework to discriminate normal control(NC) subjects from subjects with AD pathology (AD and NC, MCI subjects convert to AD in future). Our novel approach utilizing a multimodal and multiscale deep neural network was found to deliver a 85.68\% accuracy in the prediction of subjects within 3 years to conversion. Cross validation experiments proved that it has better discrimination ability compared with results in existing published literature.
\end{abstract}
\begin{document}

\flushbottom
\maketitle
%
%
\thispagestyle{empty}


\section*{Introduction}
Alzheimer's disease (AD), the most common dementia, affecting 1 out of 9 people over the age of 65 years \cite{alzheimer20112011}. Alzheimer's diseases involves progressive cognitive impairment, commonly associated with early memory loss, requiring assistance for activities of self care during advanced stages. Alzheimer's is posited to evolve through a prodromal stage which is commonly referred to as the mild cognitive impairment(MCI) stage and 10 - 15\% of individuals with MCI, progress to AD\cite{petersen2009mild} each year. With improved life expectancy, it is estimated that about 1.2\% of global population will develop Alzheimer's disease by 2046 \cite{brookmeyer2007forecasting} thereby affecting millions of individuals directly, as well as many more indirectly through the effects on their families and caregivers. Current Alzheimer's research targets reliable prodromal identification of patients harbouring Alzheimer's pathology, for reasons that early intervention could potentially change the course of illness. Clinical diagnosis involves rigorous evaluation to rule out non-Alzheimer's causes for cognitive decline but this is however limited by specificity of identifying prodromal AD. Hence, we see the necessity of a tool to reliably detect and identify prodromal Alzheimer's disease.

Efforts to understand AD pathology in the past resulted in identifying neuroimaging as one of the promising tool for prodromal diagnosis \cite{Weiner2017}. Neuroimaging involving magnetic resonance imaging (MRI) \cite{davatzikos2011prediction} and fluorodeoxyglucose positron emission tomography (FDG-PET) \cite{landau2011associations} were the unique imaging modalities recognized as useful tools to identify individuals with prodromal AD. MRI offers structural details such as texture, thickness, density and shape of various brain regions, while FDG-PET measures the resting state glucose metabolism \cite{mosconi2010pre}, reflecting the functional activity of the underlying tissue \cite{landau2011associations}. FDG-PET and MRI are frequently employed neuroimaging techniques in computer-aided diagnosis of neurodegenerative diseases. There has been considerable efforts to use structural MRI \cite{farhan2014ensemble,korolev2017residual,payan2015predicting} or FDG-PET \cite{gray2012multi,toussaint2012resting} or a combination with other biomarkers \cite{young2013accurate,zhang2011multimodal} to develop automated computer aided tools for prodromal diagnosis of Alzheimer's disease.

Deep neural networks have been studied extensively and proven to have the best performance for many recognition tasks \cite{krizhevsky2012imagenet}. The application of deep neural networks in recognition of AD-related patterns has also attracted interests in its application for prodromal AD \cite{liu2015multimodal, liu2014early, suk2014hierarchical}. By applying deep neural network to extract features, such as stacked autoencoder(SAE) or Deep Boltzmann Machine(DBM), these approaches outperform other popular machine learning methods, e.g., support vector machine (SVM) and random forest techniques. However, one of the major hurdles preventing applications of deep learning in neuroimaging is that it requires a large data set to train the model, while the available number of images are limited to several hundred or thousand which are much less than feature dimension of data sample. To overcome this challenge, one popular approach is to segment images into patches and extract features from each patch\cite{liu2014early, liu2015multimodal, suk2014hierarchical}. However, down sampling the input data can result in the loss of discriminative information which could be a potential reason why previous methods haven't achieved satisfying accuracy for this diagnostic task. A common extension to pattern mining and feature extraction for image recognition is multiscale processing\cite{zhang2007real, lowe2004distinctive}. By extracting features at different resolutions or scales, multiscale features can better characterize images for classification task and a recent study indicates it can also improve the classification performance of deep neural networks\cite{tang2012multiresolution}.

Therefore, we are proposing a novel approach to combine multiscale and multimodal processing with deep neural network for the early diagnosis of AD. Through cross validation experiments with more than 1000 subjects, we demonstrated that 1) our network can learn hidden patterns from multiscale and multimodal features for the detection of AD pathology; 2) our approach outperform previous methods regarding the discriminative task of potential AD subjects; and, 3) our network can identify the subjects who are going to convert to AD in 3 years with an accuracy of 85.68\% which is a promising result.

\section*{Methods}
\label{methods}
There are two major steps in the proposed framework: 1)image preprocessing: segment both MRI scans and FDG-PET images into patches, and extract features from each patch; and, 2)classification: train a deep neural network to learn the patterns that discriminate AD pathology, and then use it to classify individuals with AD pathology.
\subsection*{Data}
\label{data}
Data used in the preparation of this article were obtained from the Alzheimer’s Disease Neuroimaging Initiative (ADNI) database (\url{adni.loni.usc.edu}). The ADNI was launched in 2003 as a public-private partnership, led by Principal Investigator Michael W. Weiner, MD. The primary goal of ADNI has been to test whether serial MRI, PET, other biological markers, and clinical and neuropsychological assessment can be combined to measure the progression of mild cognitive impairment (MCI) and early Alzheimer’s disease (AD). 

For a comprehensive validation of the proposed method, it is emphasized that all the available ADNI subjects(N = 1242) with both a T1-weighted MRI scan and FDG-PET image at the time of preparation of this manuscript were used in this study. These subjects were categorized into 5 groups: 1) stable Normal controls (sNC): 360 subjects diagnosed to be NC at baseline and remained the same at the time of preparation of this manuscript; 2) stable MCI (sMCI): 409 subjects diagnosed to be MCI at all time points(at least for 2 years); 3) progressive NC (pNC): 18 subjects evaluated to be NC at baseline but have progressed to probable AD at the time of preparation of this manuscript; 4) progressive MCI (pMCI): 217 subjects evaluated to be MCI at baseline and progressed to probable AD; 5) stable Alzheimer’s disease (sAD): 238 subjects diagnosed to be AD for all available time points. Demographic and clinical information of the subjects are shown in Table \ref{table:Demog}. Numbers in brackets are the number of male and female subjects in the second row,, while in the rest 3 rows the two number represent the minimum and maximum value of age, education year and MMSE(Mini–Mental State Examination) score. It was worth mentioning that each subject could have multiple scans at different time points. In total there were 2402 FDG-PET scans and 2402 MRI images used in this study. Detailed descriptions of the ADNI subject cohorts, image acquisition protocols procedures and post-acquisition preprocessing procedures can be found at \url{http://www.adni-info.org}.

\begin{table}[ht]
\centering
\setlength{\tabcolsep}{0.1in}
\begin{tabular}{c c c c c c}
\toprule
\textbf{Mean(min-max)} &\textbf{sNC} &\textbf{sMCI} &\textbf{pNC} &\textbf{pMCI}	&\textbf{sAD} \\
\midrule
\textbf{Count (M/F)} &360(167/193) &409(239/170) &18(11/7) &217(126/91) &238(141/97)\\
\textbf{Age} &73.4(60-94) &74(56-91) &77(68-84) &74(55-89) &75(55-90)\\
\textbf{Education} &16.5(6-20) &15.8(7-20) &15.7(12-20) &16.0(8-20) &15.3(4-20)\\
\textbf{MMSE} &29.1(24-30) &28.0(22-30) &29.4(27-30) &26.5(9-30) &23.2(18-27)\\
\bottomrule
\end{tabular}
\caption{Subject Demographics. In the second row, the first number in each cell is the total number of subjects and the numbers in brackets represents the number of male and female. In the last 3 rows, the first number in each cell is the mean and the two numbers in brackets are the minimum and maximum value of age, education year and MMSE score, respectively. }
\label{table:Demog}
\end{table}%

\subsection*{Image Processing}
\label{imageprocessing}
\begin{figure}[htb]
\begin{minipage}[b]{1.0\linewidth}
  \centering
  \centerline{\includegraphics[width=15cm]{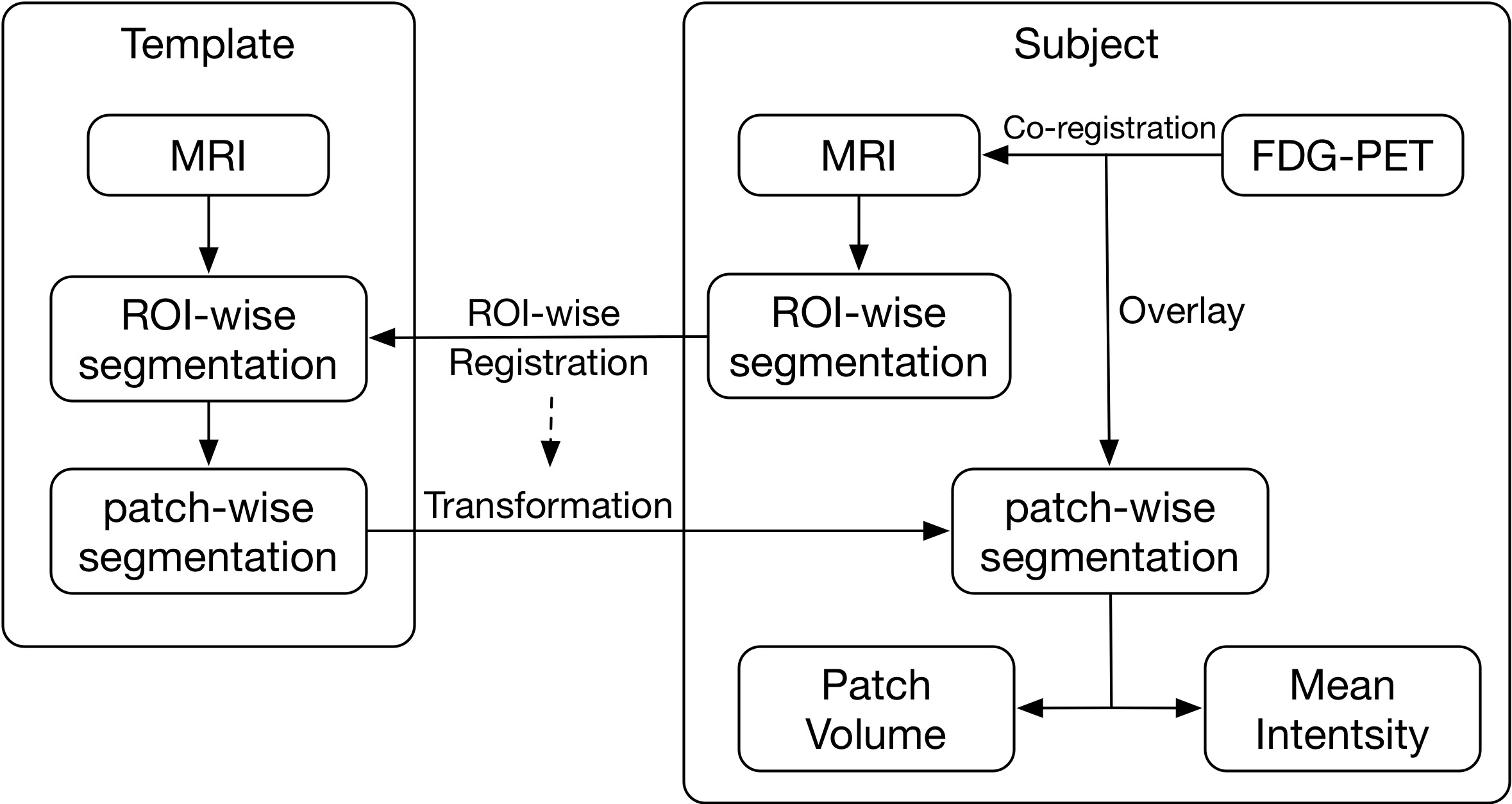}}
\end{minipage}
\caption{Flowchart of extracting patch-wise features from MRI scans and FDG-PET images. Each subject was segmented into patches through registration to a segmented template. Patch volume and mean intensity of FDG-PET were extracted as the feature to represent each patch.}
\label{fig:ImageProcessing}
\end{figure}

Unlike typical image recognition problems where deep learning has shown to be effective, our data set is relatively small. Hence directly using this smaller database of images to train the deep neural network is not likely to deliver high classification accuracy. However, contrary to typical image recognition tasks, where the images contain large heterogeneity, the images in this database are all human brain images acquired with similar pose and scale which show relatively much less heterogeneity in comparison. Therefore we applied following processing steps to extract patch-wise features as shown in Figure \ref{fig:ImageProcessing}: FreeSurfer 5.1\cite{C1} was used to segment each T1 structural MRI image into gray matter and white matter followed by subdivision of the gray matter into 87 anatomical regions of interest (ROI). The Freesurfer segmentation were quality controlled by an expert neuroanatomist and any errors noted were manually corrected. Then, for a standard T1 MRI image, a voxel-wise $k$-means clustering based on spatial coordinates was performed to segment each ROI into patches based on its spatial information\cite{raamana2015thickness}. The size of patch was predefined as 500, 1000 and 2000 voxels in this study and resulted into 1488, 705 and 343 patches, respectively. It was designed to keep enough detailed information as well as avoiding too large feature dimension considering the limited number of data samples were available in this study. Subsequently, each ROI of the standard template MRI was registered to the same ROI of every target image via a high-dimensional non-rigid registration method (LDDMM\cite{C2}). The registration maps were then applied to the patch-wise segmentation of the standard template. This transformed the template segmentation into each target MRI space so the target images were subdivided into the same number of patches. It worth mentioning that after the transformation, the size of a template patch in different images is not the same due to non-rigid registration encoding local expansion/contraction and hence is one of the features used to represent the regional information of a given structural brain scan. Then, for each target subject, the FDG-PET image of the subject was co-registered to its skull-stripped T1 MRI scan with a rigid transformation using FSL-FLIRT program\cite{jenkinson2002improved} based on normalized mutual information. The degrees of freedom (DOF) was set as $12$ and Normalized correlation was used as cost function. The mean intensity in the brainstem region of the FDG-PET image was the chosen reference to normalize the voxel intensities in that individual brain metabolism image, because brainstem region was most unlikely to be affected by AD. The mean intensity of each patch was used as an element to form the feature vector to represent the metabolism activity, and the volume of each patch was used to represent the brain atrophy.

\subsection*{Multiscale Deep Neural Network}
\label{processing}
\begin{figure}[htb]
\begin{minipage}[b]{1.0\linewidth}
  \centering
  \centerline{\includegraphics[width=15cm]{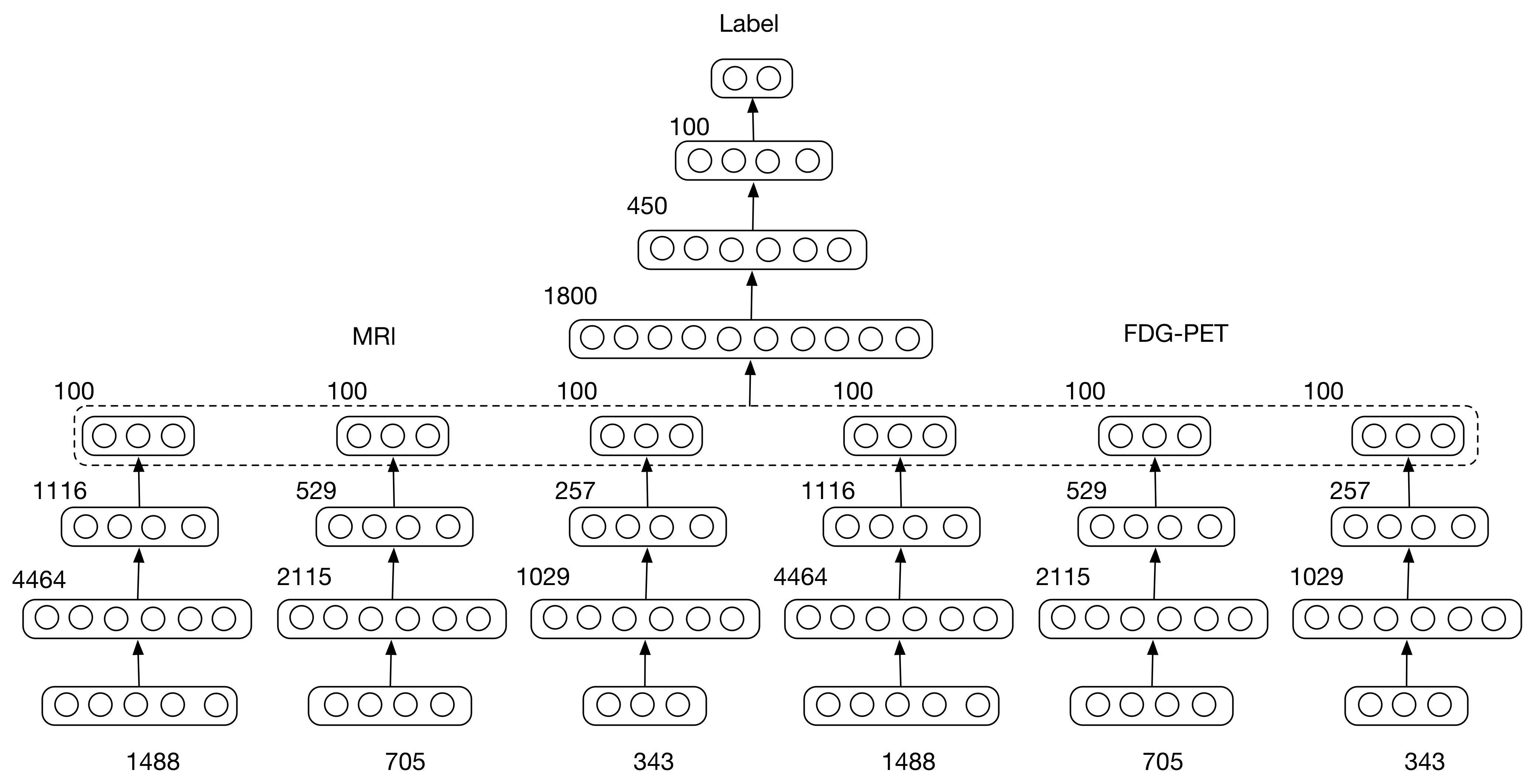}}
\end{minipage}
\caption{Multimodal Multiscale Deep Neural Network. 1488, 705 and 343 are input feature dimension(number of patches) extracted from different scales. For each layer, its number of nodes was denoted by the number on its top left.  For each scale of each image modality, its patch-wise biomarkers were feed to a single DNN. The features from these 6 DNNs were fused by another DNN to generate the final classification result.}
\label{fig:Network}
\end{figure}

With the features extracted from MRI and FDG-PET images, we trained a Multimodal Multiscale Deep Neural Network (MMDNN) to perform the classification. As shown in Fig \ref{fig:Network}, the network consists of two parts. The first parts was 6 independent deep neural network (DNN) corresponding to each scale of a single modality. The second part was another DNN used to fuse the features extracted from these 6 DNNs. The input data of this DNN was the concatenated latent representation learned from each single DNN. The DNNs in two parts shared the same structure. For each DNN, the number of nodes for each hidden layer were set as $3N$, $\frac{3}{4}N$ and 100 respectively, where $N$ denotes the dimension of input feature vector. The number of nodes was chosen to explore all possible hidden correlation across features from different patches in the first layer and gradually reduce the number of features in the following layers to avoid over-fitting. We trained each DNN with two steps, unsupervised pre-training and supervised fine-tuning, respectively. Then all the parameters of MMDNN was tuned together. 

\begin{itemize}
\item \textbf{Unsupervised Pre-training}
For unsupervised pre-training step, each DNN was trained as a stacked-autoencoder(SAE). Autoencoder is an artificial neural network used for unsupervised learning of non-linear hidden patterns from input data. It consists of three layers, input layer, hidden layer and output layer, for which two nearby layers are fully-connected. Three functions are used to define an autoencoder, encoding function, decoding function and loss function. In this study, encoding function is defined as: $y=s(W_1x+b_1)$, where $x$ is the input data, $y$ is the latent representation, $W_1$ is the weight matrix, $b_1$ is the bias term and $s$ is the activation function for which we used rectified linear function $max(0,x)$. Similarly, decoding function can be represented as: $z=s(W_2y+b_2)$, where we constrained it with tied weight $W_1=W^T$ and $z$ is the reconstructed data which is supposed to be close to input $x$. Squared error $\frac{1}{2}||x-z||^2$ is applied as loss function to optimize the network. The hypothesis is that the latent representation can capture the main factors of variation in the data. Comparing with another popular unsupervised feature learning method principle component analysis (PCA), the activation function enables the network to capture non-linear factors of data variation, especially when multiple encoders and decoders are stacked to form a SAE. To fully train the network, we applied greedy layer wise training\cite{bengio2007greedy} approach where every hidden layer was trained separately.

\item \textbf{Supervised Fine-tuning}
After pre-training, the first three layers of a DNN were initialized with the parameters of encoders from pre-trained SAE followed by a softmax output layer. At first we trained the output layer independently while fixing the parameters of first 3 layers. Then we fine-tuned the whole network as Multilayer Perceptron (MLP) with subject labels for criterion. The network outputs the probabilities of a subject belongs to each class and the class with highest probability determines the output label of the subject. If we use $x^i$ , $y^i$ to represent the input feature vector and label of the $i_{th}$ sample, respectively, the loss function based on cross entropy can be displayed as:
\begin{equation}
\quad\quad\quad\quad\quad\quad\quad\quad\quad\quad\quad\quad\quad\quad
H(i)=-\frac{1}{N}\sum_{i=1}^{N}\sum_{j=1}^{2}[\mathbbm{1}\{y^i=j\}log(h(x^i)_j]
\end{equation}
where $N$ is the number of input samples, $j$ represents the class of samples, and $h$ represents the network function.

\item \textbf{Optimization of Network} Every training step of the networks were performed via back propagation with Adam algorithm\cite{kingma2014adam}. It is a first-order gradient-based optimization algorithm which has been proven to be computational efficient and appropriate for training deep neural network. During training stage, the training set was randomly split into mini batches\cite{bengio2012practical} where each of them contains 50 samples in this study. At every iteration, only a single mini batch was used for optimization. After every batch has been used once, the training set was reordered and randomly divided again so that each batch would have different samples in different epoch.

\item \textbf{Dropout}
In order to prevent deep neural network from overfitting, regularization is necessary to reduce its generalization error. In this study, we used dropout\cite{srivastava2014dropout} to learn more robust features and prevent overfitting. In the dropout layer, some units were randomly dropped, providing a way to combine exponentially many different neural networks.  In this study, we inserted dropout layers after every hidden layer. In each iteration of training stage, only half of hidden units were randomly selected to feed the results to the next layer, while in the testing stage all hidden units were kept to perform the classification. By avoiding training all hidden units on every training sample, this regularization technique not only prevented complex co-adaptations on training data and decrease overfitting, but also reduced the amount of computation and improved training speed. 

\item \textbf{Early Stopping}
Another approach we used to prevent overfitting is early stopping. Because deep architecture were trained with iterative back propagation, the network were prone to be more adaptive to training data after every epoch. At a certain point, improving the network's fit to the training set will start to decrease generalization accuracy. In order to terminate the optimization algorithm before over-fitting, early stopping was used to provide guidance for how many iterations are needed. In the cross validation experiment, after dividing the data set into training and testing, we further split the training samples into a training set and a validation set. The networks were trained only with data in the former set, while samples in the latter set was used to determine when to stop the algorithm: while the network has the highest generalization accuracy for validation set. In actual training, we stopped the optimization if the validation accuracy had ceased to increase for 50 epochs.

\item \textbf{Ensemble Classifiers}
Although early stopping has proven to be useful in most deep learning problems, relatively small data set limited the number of samples we could use for validation. And a small validation set may not able to represent the whole data set resulting in a biased network. Therefore, we resorted to ensemble multiple classifiers to perform more stable and robust classification. Instead of selecting a single validation set, we randomly divided the training set into 10 sets and used them to train 10 different networks to 'vote' for the classification. At the training stage, for network $i$, set $i$ would be used for validation while the rest 9 sets were used for training. At the testing stage, the test samples were feed into all these networks resulting in 10 sets of probabilities. For each sample, the probabilities from 10 networks was added and the class with highest probability was the classification result of this sample. Although the performance of ensemble classifiers may not be better than single network in every occasion, this strategy can statistically improve the classification accuracy as well as the robustness and stability of the classifier.
\end{itemize}

\section*{Results and Discussion}
\label{experiments}
\subsection*{Experiments Setup}
The proposed deep neural network was built with Tensorflow\cite{tensorflow2015-whitepaper}, an open source deep learning toolbox provided by Google. First, to compare the discriminant ability with state-of-the-art methods, 10-fold cross validation experiments were applied to classify sMCI and pMCI subjects. Then we performed three experiments with different training sets to test whether the images of pNC and pMCI contain AD pathology or not. For these experiments, 4 data sets: sNC, sAD, pNC and pMCI, were divided into two groups in 3 different ways: 1) subjects of sNC and sAD were considered as group 1, and subjects of pNC and pMCI belonged to group 2; 2) subjects of pMCI, sNC and sAD belonged to group1, and pNC were considered as group2; 3) all subjects were considered as group1. For each experiment, we applied a 10-fold cross validation on group1. The subjects of group 1 were randomly divided into 10 subsets, with 9 sets used for training and the rest set combined with subjects of group 2 used for testing. As detailed in the Methods Section, 10\% of training subjects were randomly selected as validation set for early stopping to prevent overfitting. 10 networks with different validation set were trained to 'vote' for the final classification result. Noticing it was not images but subjects we were splitting, so images from different time point of the same subject won't be used for both training and testing. 

\subsection*{Compare with State-of-the-Art Methods}
Researchers in the past have worked on classification of subjects with progressive cognitive decline and those with stable cognitive impairment. pMCI were recognized as individuals with high risk of AD, while the sMCI were considered as those with no risk or low risk of AD in these studies. To evaluate the performance of our approach, we compared the classification accuracy of pMCI vs. sMCI with previous methods using the same data modality, i.e. T1-weighted MRI and FDG-PET \cite{young2013accurate,liu2017prediction,cheng2015domain,suk2014hierarchical}. The proposed network outperformed the state-of-the-art method in classifying pMCI and sMCI individuals, irrespective of using single or multimodal imaging, as shown in Table \ref{table:sMCIvspMCI}. It is worth to mention that in the study of Chen et.al \cite{cheng2015domain}, they performed domain transfer learning to exploit the auxiliary domain data(AD/NC subjects) to improve the classification. Even though, the acuccracy of our methods without auxiliary knowledge was 3.5\% accuracy than theirs.

\begin{table}[h]
\centering
\setlength{\tabcolsep}{0.1in}
\begin{center}
\begin{tabular}{cccccc}
\multicolumn{1}{c}{\bf Method} &\multicolumn{1}{c}{\bf Modality} &\multicolumn{1}{c}{\bf \# Subjects} &\multicolumn{1}{c}{\bf Accuracy} &\multicolumn{1}{c}{\bf Sensitivity} &\multicolumn{1}{c}{\bf Specificity}
\\ \hline
Young et al. &MRI &143 &64.3 &53.2 &69.8\\
Liu et al. &MRI &234 &68.8 &64.29 &74.07 \\
Suk et al. &MRI &204 &72.42 &36.7 &90.98 \\
Cheng et al. &MRI &99 &73.4 &74.3 &72.1\\
\textbf{Proposed} &MRI &\textbf{626} &\textbf{75.44} &77.27 &76.19\\
Young et al. &PET &143 &65.0 &66.0 &64.6\\
Liu et al. &PET &234 &68.8 &57.14 &82.41\\
Suk et al. &PET &204 &70.75 &25.45 &96.55\\
Cheng et al. &PET &99 &71.6 &76.4 &67.9\\
\textbf{Proposed} &PET &\textbf{626} &\textbf{81.53} &78.20 &82.47\\
Young et al. &MRI+PET+APOE &143 &69.9 &78.7 &65.6\\
Liu et al. &PET+MRI &234 &73.5 &76.19 &70.37 \\
Suk et al. &PET+MRI &204 &75.92 &48.04 &95.23\\
Cheng et al. &PET+MRI+CSF &99 &79.4 &84.5 &72.7\\
\textbf{Proposed} &PET+MRI &\textbf{626} &\textbf{82.93} &79.69 &83.84\\
\hline
\end{tabular}
\end{center}
\caption{Accuracy(\%), Sensitivity(\%), and Specificity(\%) of the proposed network comparing with state-of-the-art methods. The third row is the number of subjects used in the experiments}
\label{table:sMCIvspMCI}
\end{table}

\subsection*{AD Pathology Classification}
\begin{figure*}[tbh!]
\centering
\begin{center}
\includegraphics[width=17cm,height=4cm]{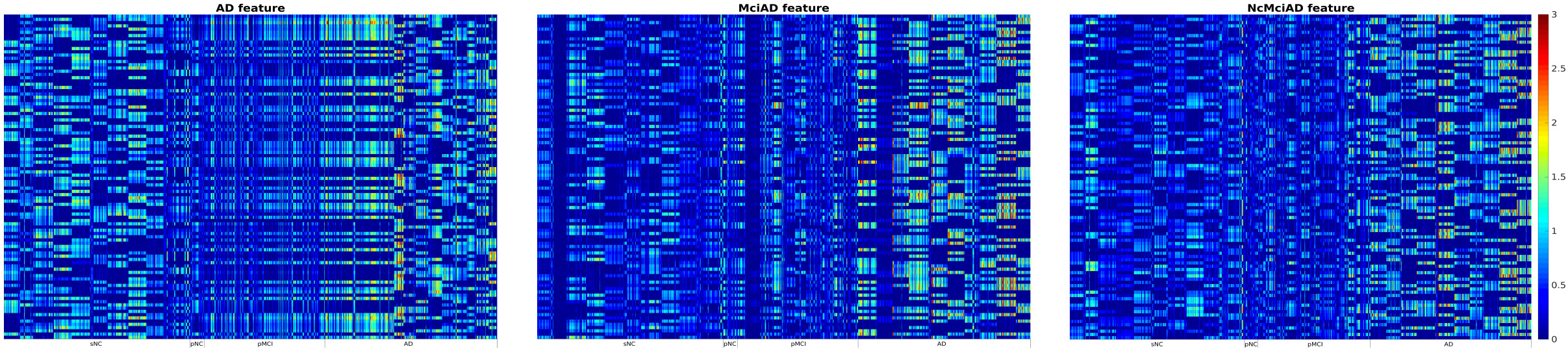}
\end{center}
\caption{Features fed to the output layer. From left to right the training set is sNC vs sAD, sNC vs sAD and pMCI, and sNC vs sAD, pMCI and pNC, respectively. The y axis represents the units of the second from last layer, while x axis denotes the different data samples. This figure shows the different patterns as distilled by the deep learning network from the sNC, pNC, pMCI and AD images.}
\label{fig:feature}
\end{figure*}

One problem of sMCI subjects was that we only know they remained stable at the time of preparation this manuscript, but they could still progress to AD or other mental disease in the future. Although the the sMCI vs. pMCI experiment were commonly used to test the discriminate ability of classifiers in recent studies, the classification result of sMCI subjects may not be very accurate. Therefore, we performed the second experiment involved classifying individuals with only known Alzheimer's progression (pNC, pMCI and sAD) and normal controls (sNC). We investigated the performance of the classifier by using various combinations of samples during training phase. At a very basic level, we trained the classifier by discriminating sAD and sNC, at the next level pMCI and sAD were combined to represent the Alzheimer's group and trained to discriminate them from the sNC group. In the last level we combined pNC, pMCI and sAD to represent the Alzheimer's group to discriminate from the sNC group. The features extracted by the deep neural network are displayed in Fig \ref{fig:feature}. We observed the accuracy and sensitivity of the classifier progressively improved by additionally training with pMCI and pNC, while the specificity decreased, as displayed in Table \ref{table:result}. Further, the classifier performance was marginally better with the combination of FDG-PET and structural measurements compared to the performances with individual modalities. Interestingly, the classifier performance of structural imaging measurements were inferior to that of FDG-PET measurements. Supporting, the fact that FDG-PET, a measure of neuronal activity is a better tool to identify prodromal Alzheimer's as compared to structural images \cite{Kawachi2006ComparePETandMRI, Chetelat2008DirectComparisonAtrophyHypometabolism}. 

\subsection*{Multiscale Classification}
The classification accuracy of features extracted with different scales are listed in Table \ref{table:patch}. We could not recognize any trend of increasing or decreasing classification performance with the changes in patch size. Therefore, features with higher resolution do not necessarily cover the discriminative information of lower resolution features. However, fusing multiscale features yielded superior accuracy as compared to uniscale, suggesting the network has the ability to capture discriminative information across the coarse to fine resolutions.

\begin{table}[h]
\centering
\setlength{\tabcolsep}{0.1in}
\begin{center}
\begin{tabular}{cccccc}
\multicolumn{1}{c}{\bf Training Set} &\multicolumn{1}{c}{\bf Modality} &\multicolumn{1}{c}{\bf 500}  &\multicolumn{1}{c}{\bf 1000} &\multicolumn{1}{c}{\bf 2000} &\multicolumn{1}{c}{\bf Multiscale}
\\ \hline 
sAD vs. sNC &FDG-PET &84.29 &83.76 &83.89 &84.46\\
    &MRI &81.27 &81.58 &81.01 &81.89\\
sAD and pMCI  &FDG-PET &85.34 &84.80 &84.87 &85.46\\
vs. sNC   &MRI &82.18 &82.69 &82.10 &82.77\\
sAD, pMCI  &FDG-PET &85.43 &85.28 &84.93 &85.89\\
and pNC vs. sNC   &MRI &81.69 &82.04 &81.64 &82.45\\
\hline
\end{tabular}
\end{center}
\caption{Accuracy(\%) using features at different scales of different modality.}
\label{table:patch}
\end{table}

\begin{table}[h]
\centering
\setlength{\tabcolsep}{0.1in}
\begin{center}
\begin{tabular}{c ccc ccc ccc}
 \toprule
  & \multicolumn{3}{c}{\bfseries FDG-PET} & \multicolumn{3}{c}{\bfseries Volume} & \multicolumn{3}{c}{\bfseries Multimodal}\\
  \cmidrule(lr){2-4}\cmidrule(lr){5-7}\cmidrule(lr){8-10}
  \textbf{Training Set} & Acc & Sen & Spe & Acc & Sen & Spe & Acc & Sen & Spe\\
  \hline
  sAD vs. sNC &84.46 &79.89 &91.90 &81.89 &75.49 &92.30 &84.59 &80.17 &91.77\\
  sAD and pMCI vs. sNC &85.46 &85.01 &86.19 &82.77 &79.76 &87.65 &85.96 &85.65 &86.45\\
  sAD, pMCI and pNC vs. sNC &85.89 &85.62 &86.32 &82.45 &80.23 &86.06 &86.44 &86.52 &86.32\\
  \midrule
\end{tabular}
\end{center}
\caption{Accuracy(\%), sensitivity(\%) and specificity(\%) of different modality using different training sets.}
\label{table:result}
\end{table}

\subsection*{Early Diagnosis}
We also investigated the classifier's ability to identify individuals with high risk of acquiring Alzheimer's, prior to disease onset. The classifier trained with the combined sample of Alzheimer's trajectory (pNC, pMCI and sAD) was superior, as compared to the classifier trained with sAD alone. As the network classifier was trained with patterns of AD trajectory using pNC and pMCI, the network was able to achieve exceptional classification performances in identifying individuals with AD risk, i.e the classifier recognized individuals with AD risk with a of 90.08\%, 85.61\% and 81.20\%, approximately at 1, 2, and 3 years prior to disease onset respectively. Studies in the past have predicted AD onset, using unimodal or multimodal investigation. Few studies have used PET as a single modality or in combination with structural MRI, CSF or cognitive measures to predict AD onset\cite{cheng2015domain, xu2015multi, cabral2015predicting, young2013accurate, Zhang2012, Shaffer2013}. The accuracy of 3 year prediction in the present network analysis was superior than those reported in the quoted studies. Studies predicting the illness onset, using structural MRI as a standalone tool or in addition to other clinical variables, have reported accuracy values inferior than to those obtained using PET \cite{Eskildsen2013, moradi2015machine, korolev2016predicting, Misra2009, davatzikos2011prediction, Ye2012, Gaser2013, Cuingnet2011, Wolz2011, Chupin2009, Cho2012, Coupe2012}. 

\begin{figure*}[tbh!]
\centering
\begin{center}
\includegraphics[width=17cm,height=4cm]{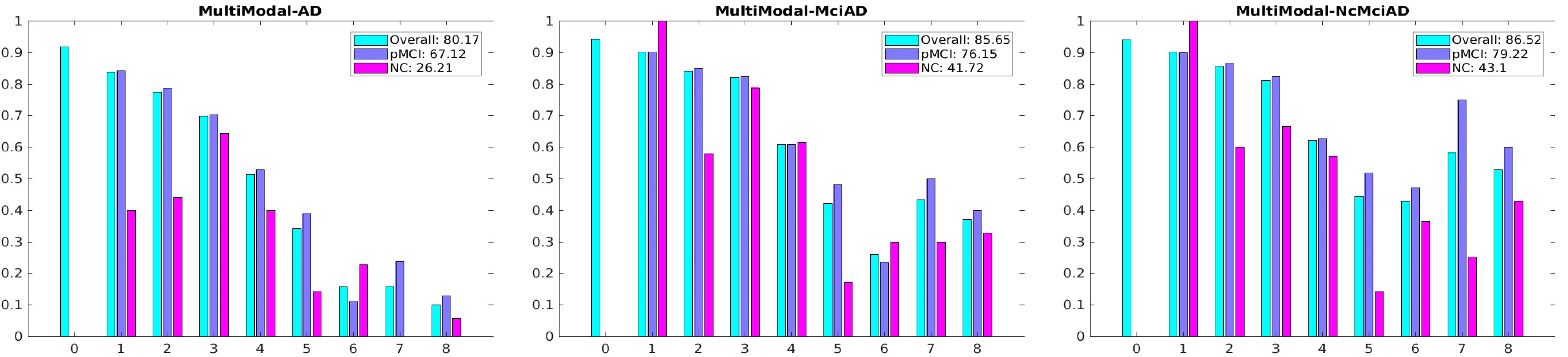}
\end{center}
\caption{Classification accuracy of different training sets. From left to right the training set is sNC vs sAD, sAD and pMCI, sAD, pMCI and pNC, respectively. The y axis represents accuracy, while x axis denotes time(year) to conversion in which '0' means the subjects are already diagnosed as AD at the imaging visit. The number in legend is the classification accuracy of all time points. Noticing the accuracy increases with more data used for training.}
\label{fig:longitudinal}
\end{figure*}

Deep neural network is a strong tool for accurate recognition of objects, by a-priori training of images with well characterized objects. Hence the basic requirement for accurate classification using DNN tool are providing large number of images (usually in millions) and well characterized objects during supervised training phase \cite{krizhevsky2012imagenet}. Therefore a compromise in the a-priori knowledge of the objects (features of Alzheimer's) provided during training would limit the accuracy of the subsequent classification. As our current understanding of AD pathogenesis and its precise characteristics in FDG-PET and structural MRI images are limited, DNN suffered jeopardy in achieving 100\% accurate classification. The clinical criteria for the diagnosis of AD involves a series of evaluations to provide near precise diagnosis. Despite rigorous evaluations, clinically diagnosed individuals with AD are not 100\% accurate and hence the FDG-PET and structural MRI characters can overlap with conditions other than AD, including NC. Therefore the DNN trained with less accurately characterized images (FDG-PET and structural MRI) was unable to achieve 100\% accurate classification. We propose an improvement in characterization of AD features by either upgrading FDG-PET and structural imaging methods or an increment in the understanding of AD specific pathogenesis, would positively impact the classification accuracy of the DNN classifier tool in future studies.

\section*{Conclusion}
In summary, we proposed a deep neural network to identify individuals at risk of developing Alzheimer's disease. We trained the classifier with patterns hidden in different resolutions and different modalities to distinguish subjects with Alzheimer's trajectory (pNC, pMCI and sAD) and those without cognitive deficits (sNC). Our results show the classifier's ability to successfully distinguish individuals with AD pathology from sNC with a remarkable accuracy of 82.93\% using cross validation experiments. We observed the performance of network classifier built by the combination of FDG-PET and structural MRI images was better than those built with either structural MRI or FDG-PET alone. Further the classifier trained with the combined sample of pNC, pMCI and sAD (Alzheimer's trajectory) was found to yield the highest classification accuracy. Lastly, our experiment to recognize individuals with AD pathology, prior to illness onset demonstrated a sensitivity of 85.68\% in 3 years earlier to illness onset. Hence, the proposed deep neural network classifier can be a potential tool of choice in the future for early prediction of AD pathology. The number of pNC subjects was limited in this study resulting in a relatively low accuracy for pNC, as more data is accumulated in the future, we expect better accuracy in the prediction of NC subjects with AD pathology.

\bibliography{main}

\section*{Consortia}
\subsection*{for the Alzheimer’s Disease Neuroimaging Initiative}
\textbf{Michael Weiner\footnotemark[2], Paul Aisen\footnotemark[3], Ronald Petersen\footnotemark[4], Cliford Jack\footnotemark[5], William Jagust\footnotemark[6], John Trojanowki\footnotemark[7], Arthur Toga\footnotemark[8], Laurel Beckett\footnotemark[9], Robert Green\footnotemark[10], Andrew Saykin\footnotemark[11], John Morris\footnotemark[12], Leslie Shaw\footnotemark[12], Jefrey Kaye\footnotemark[13], Joseph Quinn\footnotemark[13], Lisa Silbert\footnotemark[13], Betty Lind\footnotemark[13], Raina Carter\footnotemark[13], Sara Dolen\footnotemark[13], Lon Schneider\footnotemark[8], Sonia Pawluczyk\footnotemark[8], Mauricio Beccera\footnotemark[8], Liberty Teodoro\footnotemark[8], Bryan Spann\footnotemark[8], James Brewer\footnotemark[14], Helen Vanderswag\footnotemark[14], Adam Fleisher\footnotemark[14], Judith Heidebrink\footnotemark[15], Joanne Lord\footnotemark[15], Sara Mason\footnotemark[5], Colleen Albers\footnotemark[5], David Knopman\footnotemark[5], Kris Johnson\footnotemark[5], Rachelle Doody\footnotemark[16], Javier Villanueva-Meyer\footnotemark[16], Munir Chowdhury\footnotemark[16], Susan Rountree\footnotemark[16], Mimi Dang\footnotemark[16], Yaakov Stern\footnotemark[17], Lawrence Honig\footnotemark[17], Karen Bell\footnotemark[17], Beau Ances\footnotemark[12], John Morris\footnotemark[12], Maria Carroll\footnotemark[12], Mary Creech\footnotemark[2], Erin Franklin\footnotemark[12], Mark Mintun\footnotemark[12], Stacy Schneider\footnotemark[12], Angela Oliver\footnotemark[12], Daniel Marson\footnotemark[18], Randall Grifth\footnotemark[18], David Clark\footnotemark[18], David Geldmacher\footnotemark[18], John Brockington\footnotemark[18], Erik Roberson\footnotemark[18], Marissa Natelson Love\footnotemark[18], Hillel Grossman\footnotemark[19], Efe Mitsis\footnotemark[19], Raj Shah\footnotemark[20], Leyla deToledo-Morrell\footnotemark[20], Ranjan Duara\footnotemark[21], Daniel Varon\footnotemark[21], Maria Greig\footnotemark[21], Peggy Roberts\footnotemark[21], Marilyn Albert\footnotemark[22], Chiadi Onyike\footnotemark[22], Daniel D’Agostino\footnotemark[22], Stephanie Kielb\footnotemark[22], James Galvin\footnotemark[23], Brittany Cerbone\footnotemark[23], Christina Michel\footnotemark[23], Dana Pogorelec\footnotemark[23], Henry Rusinek\footnotemark[23], Mony de Leon\footnotemark[23], Lidia Glodzik\footnotemark[23], Susan De Santi\footnotemark[23], P. Doraiswamy\footnotemark[24], Jefrey Petrella\footnotemark[24], Salvador Borges-Neto\footnotemark[24], Terence Wong\footnotemark[24], Edward Coleman\footnotemark[24], Charles Smith\footnotemark[25], Greg Jicha\footnotemark[25], Peter Hardy\footnotemark[25], Partha Sinha\footnotemark[25], Elizabeth Oates\footnotemark[25], Gary Conrad\footnotemark[25], Anton Porsteinsson\footnotemark[26], Bonnie Goldstein\footnotemark[26], Kim Martin\footnotemark[26], Kelly Makino\footnotemark[26], M. Ismail\footnotemark[26], Connie Brand\footnotemark[26], Ruth Mulnard\footnotemark[27], Gaby Thai\footnotemark[27], Catherine Mc-Adams-Ortiz\footnotemark[27], Kyle Womack\footnotemark[28], Dana Mathews\footnotemark[28], Mary Quiceno\footnotemark[28], Allan Levey\footnotemark[29], James Lah\footnotemark[29], Janet Cellar\footnotemark[29], Jefrey Burns\footnotemark[30], Russell Swerdlow\footnotemark[30], William Brooks\footnotemark[30], Liana Apostolova\footnotemark[31], Kathleen Tingus\footnotemark[31], Ellen Woo\footnotemark[31], Daniel Silverman\footnotemark[31], Po Lu\footnotemark[31], George Bartzokis\footnotemark[31], Neill Graf-Radford\footnotemark[32], Francine Parftt\footnotemark[32], Tracy Kendall\footnotemark[32], Heather Johnson\footnotemark[32], Martin Farlow\footnotemark[11], Ann Marie Hake\footnotemark[11], Brandy Matthews\footnotemark[11], Jared Brosch\footnotemark[11], Scott Herring\footnotemark[11], Cynthia Hunt\footnotemark[11], Christopher Dyck\footnotemark[33], Richard Carson\footnotemark[33], Martha MacAvoy\footnotemark[33], Pradeep Varma\footnotemark[33], Howard Chertkow\footnotemark[34], Howard Bergman\footnotemark[34], Chris Hosein\footnotemark[34], Sandra Black\footnotemark[35], Bojana Stefanovic\footnotemark[35], Curtis Caldwell\footnotemark[35], Ging-Yuek Robin Hsiung\footnotemark[36], Howard Feldman\footnotemark[36], Benita Mudge\footnotemark[36], Michele Assaly\footnotemark[36], Elizabeth Finger\footnotemark[37], Stephen Pasternack\footnotemark[37], Irina Rachisky\footnotemark[37], Dick Trost\footnotemark[37], Andrew Kertesz\footnotemark[37], Charles Bernick\footnotemark[38], Donna Munic\footnotemark[38], Marek-Marsel Mesulam\footnotemark[39], Kristine Lipowski\footnotemark[39], Sandra Weintraub\footnotemark[39], Borna Bonakdarpour\footnotemark[39], Diana Kerwin\footnotemark[39], Chuang-Kuo Wu\footnotemark[39], Nancy Johnson\footnotemark[39], Carl Sadowsky\footnotemark[40], Teresa Villena\footnotemark[40], Raymond Scott Turner\footnotemark[41], Kathleen Johnson\footnotemark[41], Brigid Reynolds\footnotemark[41], Reisa Sperling\footnotemark[42], Keith Johnson\footnotemark[42], Gad Marshall\footnotemark[42], Jerome Yesavage\footnotemark[43], Joy Taylor\footnotemark[43], Barton Lane\footnotemark[43], Allyson Rosen\footnotemark[43], Jared Tinklenberg\footnotemark[43], Marwan Sabbagh\footnotemark[44], Christine Belden\footnotemark[44], Sandra Jacobson\footnotemark[44], Sherye Sirrel\footnotemark[44], Neil Kowall\footnotemark[45], Ronald Killiany\footnotemark[45], Andrew Budson\footnotemark[45], Alexander Norbash\footnotemark[45], Patricia Lynn Johnson\footnotemark[45], Thomas Obisesan\footnotemark[46], Saba Wolday\footnotemark[46], Joanne Allard\footnotemark[46], Alan Lerner\footnotemark[47], Paula Ogrocki\footnotemark[47], Curtis Tatsuoka\footnotemark[47], Parianne Fatica\footnotemark[47], Evan Fletcher\footnotemark[48], Pauline Maillard\footnotemark[48], John Olichney\footnotemark[48], Charles DeCarli\footnotemark[48], Owen Carmichael\footnotemark[48], Smita Kittur\footnotemark[49], Michael Borrie\footnotemark[50], T-Y Lee\footnotemark[50], RobBartha\footnotemark[50], Sterling Johnson\footnotemark[51], Sanjay Asthana\footnotemark[51], Cynthia Carlsson\footnotemark[51], Steven Potkin\footnotemark[52], Adrian Preda\footnotemark[52], Dana Nguyen\footnotemark[52], Pierre Tariot\footnotemark[53], Anna Burke\footnotemark[53], Nadira Trncic\footnotemark[53], Adam Fleisher\footnotemark[53], Stephanie Reeder\footnotemark[53], Vernice Bates\footnotemark[54], Horacio Capote\footnotemark[54], Michelle Rainka\footnotemark[54], Douglas Scharre\footnotemark[55], Maria Kataki\footnotemark[55], Anahita Adeli\footnotemark[55], Earl Zimmerman\footnotemark[56], Dzintra Celmins\footnotemark[56], Alice Brown\footnotemark[56], Godfrey Pearlson\footnotemark[57], Karen Blank\footnotemark[57], Karen Anderson\footnotemark[57], Laura Flashman\footnotemark[58], Marc Seltzer\footnotemark[58], Mary Hynes\footnotemark[58], Robert Santulli\footnotemark[58], Kaycee Sink\footnotemark[59], Leslie Gordineer\footnotemark[59], Jef Williamson\footnotemark[59], Pradeep Garg\footnotemark[59], Franklin Watkins\footnotemark[59], Brian Ott\footnotemark[60], Henry Querfurth\footnotemark[60], Geofrey Tremont\footnotemark[60], Stephen Salloway\footnotemark[61], Paul Malloy\footnotemark[61], Stephen Correia\footnotemark[61], Howard Rosen\footnotemark[62], Bruce Miller\footnotemark[62], David Perry\footnotemark[62], Jacobo Mintzer\footnotemark[63], Kenneth Spicer\footnotemark[63], David Bachman\footnotemark[63], Nunzio Pomara\footnotemark[64], Raymundo Hernando\footnotemark[65], Antero Sarrael\footnotemark[64], Norman Relkin\footnotemark[65], Gloria Chaing\footnotemark[65], Michael Lin\footnotemark[65], Lisa Ravdin\footnotemark[65], Amanda Smith\footnotemark[66], Balebail Ashok Raj\footnotemark[66] \& Kristin Fargher\footnotemark[66].}

\noindent\textsuperscript{2}{Magnetic Resonance Unit at the VA Medical Center and Radiology, Medicine, Psychiatry and Neurology, University of California, San Francisco, USA.} \textsuperscript{3}{San Diego School of Medicine, University of California, California, USA.} \textsuperscript{4}{Mayo Clinic, Minnesota, USA.} \textsuperscript{5}{Mayo Clinic, Rochester, USA.} \textsuperscript{6}{University of California, Berkeley, USA.} \textsuperscript{7}{University of Pennsylvania, Pennsylvania, USA.} \textsuperscript{8}{University of Southern California, California, USA.} \textsuperscript{9}{University of California, Davis, California, USA.} \textsuperscript{10}{MPH Brigham and Women’s Hospital/Harvard Medical School, Massachusetts, USA.} \textsuperscript{11}{Indiana University, Indiana, USA.} \textsuperscript{12}{Washington University St. Louis, Missouri, USA.} \textsuperscript{13}{Oregon Health and Science University, Oregon, USA.} \textsuperscript{14}{University of California–San Diego, California, USA.} \textsuperscript{15}{University of Michigan, Michigan, USA.} \textsuperscript{16}{Baylor College of Medicine, Houston, State of Texas, USA.} \textsuperscript{17}{Columbia University Medical Center, South Carolina, USA.} \textsuperscript{18}{University of Alabama – Birmingham, Alabama, USA.} \textsuperscript{19}{Mount Sinai School of Medicine, New York, USA.} \textsuperscript{20}{Rush University Medical Center, Rush University, Illinois, USA.} \textsuperscript{21}{Wien Center, Florida, USA.} \textsuperscript{22}{Johns Hopkins University, Maryland, USA.} \textsuperscript{23}{New York University, NY, USA.} \textsuperscript{24}{Duke University Medical Center, North Carolina, USA.} \textsuperscript{25}{University of Kentucky, Kentucky, USA.} \textsuperscript{26}{University of Rochester Medical Center, NY, USA.} \textsuperscript{27}{University of California, Irvine, California, USA.} \textsuperscript{28}{University of Texas Southwestern Medical School, Texas, USA.} \textsuperscript{29}{Emory University, Georgia, USA.} \textsuperscript{30}{University of Kansas, Medical Center, Kansas, USA.} \textsuperscript{31}{University of California, Los Angeles, California, USA.} \textsuperscript{32}{Mayo Clinic, Jacksonville, Jacksonville, USA.} \textsuperscript{33}{Yale University School of Medicine, Connecticut, USA.} \textsuperscript{34}{McGill University, Montreal-Jewish General Hospital, Montreal, Canada.} \textsuperscript{35}{Sunnybrook Health Sciences, Ontario, USA.} \textsuperscript{36}{U.B.C. Clinic for AD \& Related Disorders, Vancouver, BC, Canada.} \textsuperscript{37}{Cognitive Neurology - St. Joseph’s, Ontario, USA.} \textsuperscript{38}{Cleveland Clinic Lou Ruvo Center for Brain Health, Ohio, USA.} \textsuperscript{39}{Northwestern University, San Francisco, USA.} \textsuperscript{40}{Premiere Research Inst (Palm Beach Neurology), west Palm Beach, USA.} \textsuperscript{41}{Georgetown University Medical Center, Washington DC, USA.} \textsuperscript{42}{Brigham and Women’s Hospital, Massachusetts, USA.} \textsuperscript{43}{Stanford University, California, USA.} \textsuperscript{44}{Banner Sun Health Research Institute, Sun City, AZ 85351, USA.} \textsuperscript{45}{Boston University, Massachusetts, USA.} \textsuperscript{46}{Howard University, Washington DC, USA.} \textsuperscript{47}{Case Western Reserve University, Ohio, USA.} \textsuperscript{48}{University of California, Davis – Sacramento, California, USA.} \textsuperscript{49}{Neurological Care of CNY, Liverpool, NY 13088, USA.} \textsuperscript{50}{Parkwood Hospital, Pennsylvania, USA.} \textsuperscript{51}{University of Wisconsin, Wisconsin, USA.} \textsuperscript{52}{University of California, Irvine – BIC, USA.} \textsuperscript{53}{Banner Alzheimer’s Institute, Phoenix, AZ 85006, USA.} \textsuperscript{54}{Dent Neurologic Institute, NY, USA.} \textsuperscript{55}{Ohio State University, Ohio, USA.} \textsuperscript{56}{Albany Medical College, NY, USA.} \textsuperscript{57}{Hartford Hospital, Olin Neuropsychiatry Research Center, Connecticut, USA.} \textsuperscript{58}{Dartmouth-Hitchcock Medical Center, New Hampshire, USA.} \textsuperscript{59}{Wake Forest University Health Sciences, North Carolina, USA.} \textsuperscript{60}{Rhode Island Hospital, state of Rhode Island, Providence, RI 02903, USA.} \textsuperscript{61}{Butler Hospital, Providence, Rhode Island, USA.} \textsuperscript{62}{University of California, San Francisco, USA.} \textsuperscript{63}{Medical University South Carolina, Charleston, SC 29425, USA.} \textsuperscript{64}{Nathan Kline Institute, Orangeburg, New York, USA.} \textsuperscript{65}{Cornell University, Ithaca, New York, USA.} \textsuperscript{66}{USF Health Byrd Alzheimer’s Institute, University of South Florida, Tampa, FL 33613, USA.}

\section*{Acknowledgements}
This work was supported by National Science Engineering Research Council (NSERC), Canadian Institutes of Health Research (CIHR), Michael Smith Foundation for Health Research (MSFHR), Brain Canada, Genome BC and the Pacific Alzheimer Research Foundation (PARF). 
Data collection and sharing for this project was funded by the Alzheimer's Disease Neuroimaging Initiative (ADNI) (National Institutes of Health Grant U01 AG024904) and DOD ADNI (Department of Defense award number W81XWH-12-2-0012). ADNI is funded by the National Institute on Aging, the National Institute of Biomedical Imaging and Bioengineering, and through generous contributions from the following: AbbVie, Alzheimer’s Association; Alzheimer’s Drug Discovery Foundation; Araclon Biotech; BioClinica, Inc.; Biogen; Bristol-Myers Squibb Company; CereSpir, Inc.; Cogstate; Eisai Inc.; Elan Pharmaceuticals, Inc.; Eli Lilly and Company; EuroImmun; F. Hoffmann-La Roche Ltd and its affiliated company Genentech, Inc.; Fujirebio; GE Healthcare; IXICO Ltd.; Janssen Alzheimer Immunotherapy Research \& Development, LLC.; Johnson \& Johnson Pharmaceutical Research \& Development LLC.; Lumosity; Lundbeck; Merck \& Co., Inc.; Meso Scale Diagnostics, LLC.; NeuroRx Research; Neurotrack Technologies; Novartis Pharmaceuticals Corporation; Pfizer Inc.; Piramal Imaging; Servier; Takeda Pharmaceutical Company; and Transition Therapeutics. The Canadian Institutes of Health Research is providing funds to support ADNI clinical sites in Canada. Private sector contributions are facilitated by the Foundation for the National Institutes of Health (\url{www.fnih.org}). The grantee organization is the Northern California Institute for Research and Education, and the study is coordinated by the Alzheimer’s Therapeutic Research Institute at the University of Southern California. ADNI data are disseminated by the Laboratory for Neuro Imaging at the University of Southern California.

\section*{Author contributions statement}
Donghuan Lu and Weiguang Ding built the deep neural network. Donghuan Lu and Karteek Popuri processed the neuroimage data. Rakesh Balachander and Mirza Faisal Beg designed the experiments and interpreted the results. All authors reviewed the manuscript.

\section*{Additional information}
\textbf{Competing Interest:} The authors declare that they have no competing interest.
\end{document}